\pdfoutput=1
\ifcsname UseRawInputEncoding\endcsname\UseRawInputEncoding\fi
\documentclass[runningheads,a4paper]{llncs}

\usepackage{microtype}
\usepackage{amssymb}
\usepackage{graphicx}
\usepackage{pdfpages}
\usepackage{url}
\usepackage{wrapfig}

\usepackage[colorlinks=true,linkcolor=black,citecolor=black,urlcolor=black]{hyperref}

\usepackage{eso-pic}

\AtBeginDocument{\AddToShipoutPictureFG*{\AtTextUpperLeft{\put(0,\LenToUnit{40pt}){\parbox{\textwidth}{\centering\bfseries
Proceedings of 17th RoboCup International Symposium,\\Eindhoven, Netherlands, 2013
}}}}}
\begin{document}

\mainmatter  %

\title{Humanoid TeenSize Open Platform NimbRo-OP}
\author{Max Schwarz, Julio Pastrana, Philipp Allgeuer, Michael Schreiber, Sebastian Schueller,  Marcell Missura and Sven Behnke}

\titlerunning{Humanoid TeenSize Open Platform NimbRo-OP}

\authorrunning{Schwarz \and Pastrana \and Allgeuer \and Schreiber \and Schueller  \and  Missura \and Behnke}

\institute{%
Autonomous Intelligent Systems, Computer Science, Univ.\ of Bonn, Germany\\
\{schwarzm, schuell1, behnke\}@cs.uni-bonn.de\\
\{pastrana, allgeuer, schreiber, missura\}@ais.uni-bonn.de \\
\url{http://ais.uni-bonn.de/nimbro/OP}\\[3mm]
}

\maketitle
\vspace{-5mm}
\begin{abstract}
In recent years, the introduction of affordable platforms in the
KidSize class of the Humanoid League has had a positive impact on the
performance of soccer robots. The lack of readily available larger robots, however, severely affects the number of
participants in Teen- and AdultSize and consequently the progress of research that focuses on the
challenges arising with robots of larger weight and size. This paper
presents the first hardware release of a low cost Humanoid TeenSize open
platform for research, the first software release, and the current state of ROS-based software development. 
The NimbRo-OP robot was designed to be easily manufactured, assembled, repaired, and modified. It is
equipped with a wide-angle camera, ample computing power, and enough torque to enable
 full-body motions, such as dynamic bipedal locomotion, kicking, and getting up.
\end{abstract}

\section{Introduction}

Low-cost and easy to maintain standardized hardware platforms, such as the
DARwIn-OP~\cite{DARwIn-OP}, have had a positive impact on the performance of teams in the KidSize
class of the RoboCup Humanoid League. They lower the barrier for new teams to
enter the league and make maintaining a soccer team with the required
number of players easier. Out-of-the-box capabilities like walking and kicking allow the research groups to focus on
higher-level perceptual or behavioral skills, which increases the quality of the games and, hence, 
the attractiveness of RoboCup for visitors and media. In the competition classes with larger robots, teams 
so far are forced to participate with self-constructed robots. Naturally,
this severely affects the number of participants willing to compete, and, in consequence, the 
progress of the research that attempts to solve the challenges arising with 
robots of larger weight and size.

Inspired by the success of DARwIn-OP,  we developed a first prototype of a
TeenSize humanoid robot and released it as an open platform. Our NimbRo-OP
bipedal prototype is easy to manufacture, assemble, maintain, and modify.
The prototype can be reproduced at low cost from commonly available materials 
and standard electronic components. Moreover, the robot
is equipped with configurable actuators, sufficient sensors, and enough computational power
to ensure a considerable range of operation: from image processing, over action planning, 
to the generation and control of dynamic full-body motions. These features and the 
fact that the robot is large enough for acting in real human environments
make NimbRo-OP suitable for research in relevant areas of humanoid robotics.

\section{Related Work}
\label{chap:relatedwork}

In the KidSize class, a number of robust and affordable off-the shelf products
suitable for operation on the soccer field are available. Entry-level
construction kits including \mbox{Bioloid} \cite{Bioloid} are a very cost
effective way to enter the competitions. However, the limited capabilities of
these construction kits are an obstacle for achieving the high performance required by soccer games.

With a body height of approximately 58\,cm, the Nao robot \cite{NAO}, produced by
Aldebaran Robotics, replaced Sony's quadruped Aibo as the robot used in the 
RoboCup Standard Platform League. Nao is an attractive platform because it 
offers a rich set of features and reliable walking capabilities. However, the 
fact that it is a proprietary product is a disadvantage because one has
restricted possibilities for customizing and repairing the hardware. 

The recently introduced DARwIn-OP \cite{DARwIn-OP} is very popular in the
KidSize class. In 2011 and 2012, team DARwIn won the KidSize competitions of
RoboCup and successfully demonstrated its potential. 
DARwIn-OP has been designed to be assembled and maintained by the owner, but a
fully operational version can be ordered from \mbox{Robotis}. 
Most importantly, DARwIn-OP has been released as an
open platform. Software and construction plans are available for public
access.

The limitations of the KidSize robots lay in their size. Falling---an undesired
but inevitable consequence of walking on two legs---is a negligible problem for
small robots. Larger robots, however, can suffer severe damage as 
result of a fall. A number of commercial platforms are available with sizes
larger than 120\,cm. The most prominent examples include Honda
\mbox{Asimo}~\cite{ASIMO}, the HRP~\cite{HRP-4C} series, the Toyota Partner
Robots \cite{ToyotaPartnerRobot}, and Hubo~\cite{HUBO}. The extremely high
acquisition and maintenance costs of these robots and their lack of robustness to falls 
make them unsuitable for
use in soccer games. NimbRo-OP closes the gap between large, expensive robots
and affordable small robots. 

\section{Hardware Design}
\label{chap:hardware}

\begin{wrapfigure}{r}{0.5\textwidth}
\centering
\includegraphics[width=0.48\columnwidth]{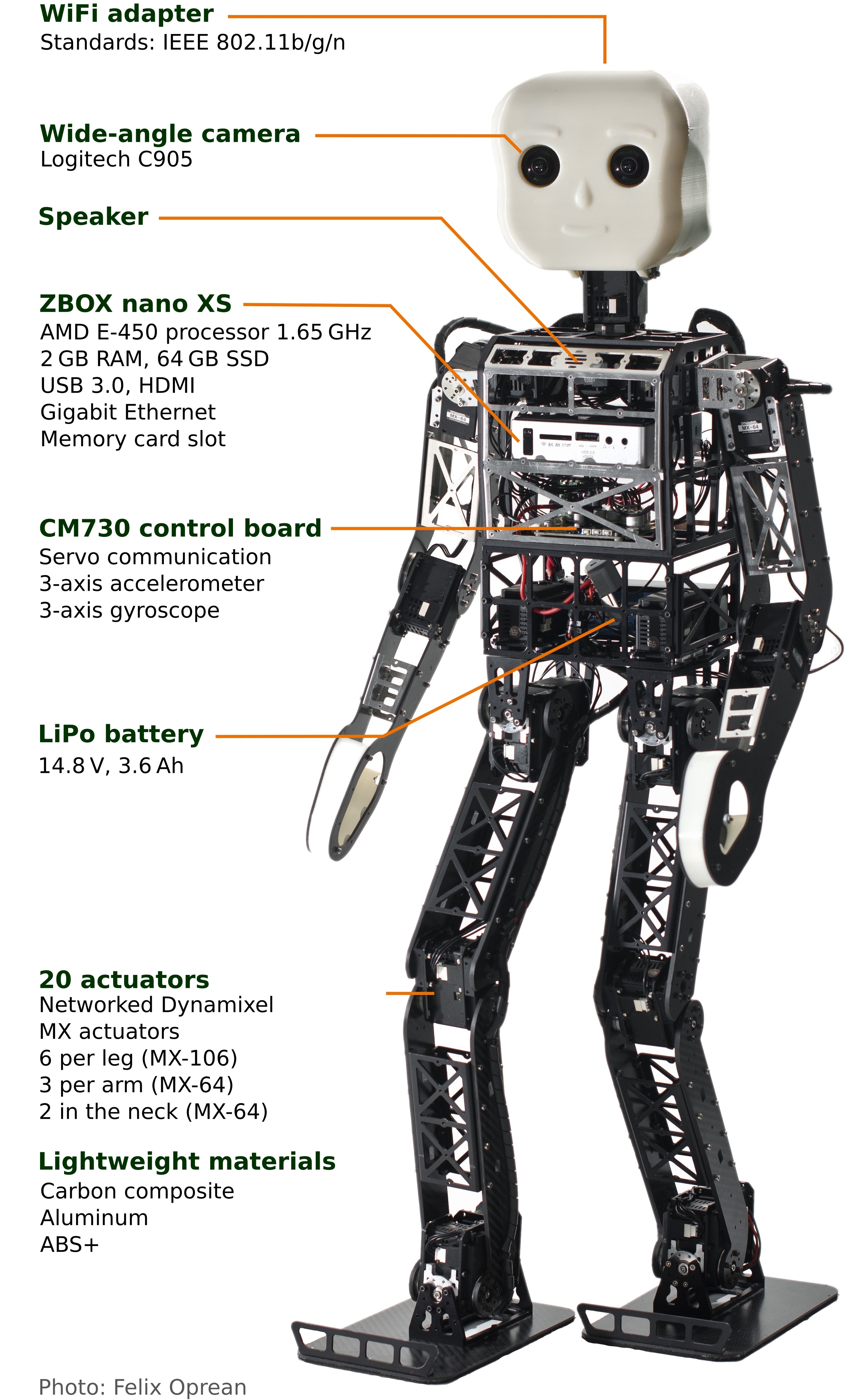}\vspace*{-1ex}
\caption{NimbRo-OP hardware.}
\label{tsop-full}
\vspace*{-4ex}
\end{wrapfigure}

The design of the NimbRo-OP hardware takes into consideration the following criteria: affordable
price, low weight, readily available parts, robot appearance, and
reproducibility in a basic workshop. Fig.~\ref{tsop-full} gives an overview of the main components.

NimbRo-OP is 95\,cm tall and weighs 6.6\,kg, including the battery. Selecting this
specific size allowed the use of a single actuator per joint. All 20 joints are driven 
by configurable Robotis Dynamixel MX 
series actuators~\cite{dynamixel-mx}. MX-106 are used in the 6\,DoF legs and MX-64 in the 3\,DoF arms and the 2\,DoF neck.  All Dynamixel
actuators are connected with a single TTL one-wire bus. The servo motors, as
well as all other electronic components are powered by a rechargeable 14.8\,V
3.6\,Ah lithium-polymer battery.

Keeping the weight as low as possible is achieved by using light-weight
materials like carbon composite and aluminum.
Arms and legs are constructed using milled carbon-composite sheets which are
connected with U-shaped aluminum parts cut from sheets and bent on two sides.
The feet are made of flexible carbon composite sheets. 
The torso is composed  of a milled rectangular aluminum
profile. The head and the connecting pieces in the hands are 3D printed using
ABS+ polymer.

NimbRo-OP is equipped with a Zotac Zbox nano XS PC featuring a dual-core
AMD E-450 1.65\,GHz processor, 2\,GB RAM, 
64\,GB solid state disk, and memory card slot. Communication interfaces are USB 3.0,
HDMI, and Gigabit Ethernet.  The PC is embedded into the torso without
modifications to facilitate upgrades. The 
head contains a USB WiFi supporting IEEE 802.11b/g/n.

A Robotis CM730 board mediates communication between the PC and the actuators.
It also contains three-axes
accelerometers and three-axes gyroscopes for attitude estimation.

The robot head contains a Logitech C905 USB camera with fish eye wide-angle lens.
Its extremely wide
field of view (ca. $180^\circ$) allows for 
simultaneously having multiple objects in sight, e.g. the ball and the goal.

The NimbRo-OP CAD files are available~\cite{github} under  
Creative Commons Attribution-NonCommercial-ShareAlike 3.0 license. This allows 
research groups to reproduce the robot and to modify it to their needs. The University of Bonn~\cite{NimbRo-OP-Website} also
offers fully assembled and tested robots.

\section{First Software Release}
\label{chap:software1}

The hardware release of the first NimbRo-OP prototype is supported by a
software package that provides a set of fundamental functionalities, such as
bipedal walking and ball tracking. Out of the box, the robot
is able to walk up to a uniformly colored ball and to kick it away. As a simple
fall protection mechanism, the robot relaxes all its joints when it detects an
inevitable fall. Moreover, the robot is able to
stand up from a prone and a supine position. We used the freely available
DARwIn-OP software framework \cite{DARwIn-OP-Software} as a
starting point for development and made only the 
modifications necessary for providing the aforementioned functionalities. More
specifically, we added \vspace*{-1ex}
\begin{itemize}
\item  correction of the wide-angle lens distortion, 
\item attitude estimation based on CM730 acceleration and turning rate sensors,
\item a feedback stabilized bipedal gait configuration,
\item an instability detection and simple fall protection mechanism, and
\item get-up and kicking motions. 
\end{itemize}
Please refer to Schwarz et al.~\cite{NimbRoOPSWS} for more details.
This software release has been made available \cite{github} as a list of small
patches against the original DARwIn-OP framework. For 
easier traceability of changes, each added or modified feature resides in an own patch 
file with a header describing the change. 

\section{ROS-based Software Development}
\label{chap:software2}

The first version of the NimbRo-OP software is based on the freely
available DARwIn-OP framework, making NimbRo-OP easily accessible to researches familiar with the 
DARwIn-OP robot, but it includes only basic skills.

\begin{figure}[tbh]
\centering
\includegraphics[trim = 138mm 36mm 65mm 26mm, clip, width=1.0\textwidth]{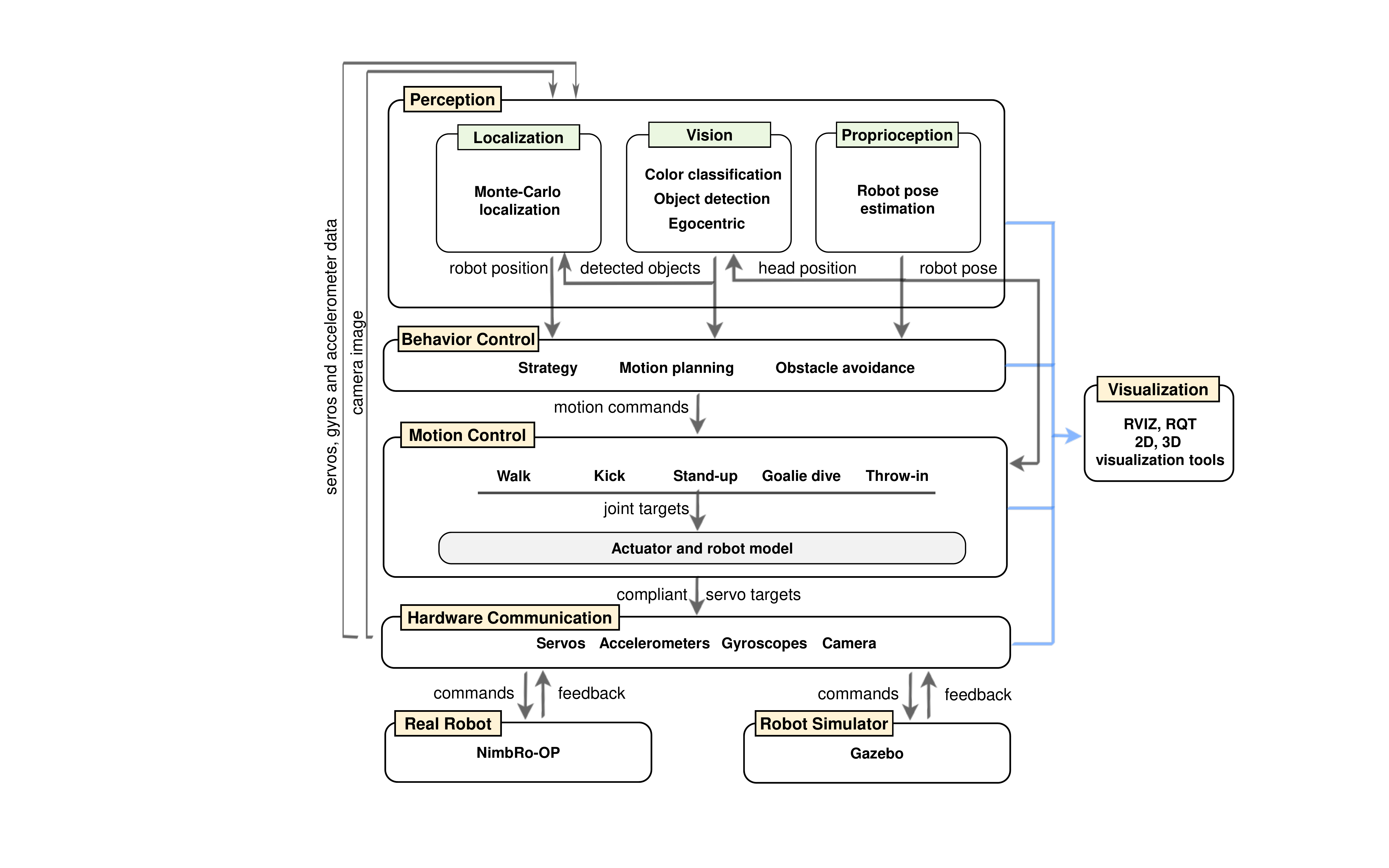}
\vspace*{-3ex}
\caption{Robot control diagram.}
\label{fig-grobotcontrolsketch}
\end{figure}

We aim for more advanced functionality, including cognitive perceptual and action capabilities. 
To this end, we started a new software development based on the Robot Operative System framework (ROS) \cite{ROS,ROSopensource}.
This popular middleware makes it easy to implement multiple
processes (nodes) that communicate with each another using data streams 
(topics)---facilitating modular software development.
Additionally, effective development tools are provided that perform tasks such
as data logging and serialization, specifying launch configurations, unit testing, and
visualization.
Most importantly, the ROS community includes thousands of robotic researchers
and enthusiasts who are constantly contributing their experiences and results.

Our ROS-based software will incorporate accumulated experience, scientific achievements, and technologies developed
throughout the successful history of team NimbRo. Fig.~\ref{fig-grobotcontrolsketch} depicts 
the initial structure of the soccer robot software that is being implemented. At the bottom of the figure 
are nodes that interface the robot and abstract from its hardware. 
The top of the figure,  Perception, is composed of 
nodes that interpret all available sensory data to estimate the robot and environment state. The Behavior Control modules, shown in the middle of the figure, decide the robot actions based on the perceived state. The Motion Control modules generate walking, kicking, and other full-body motion and send joint targets to the robot.

\subsection{Configuration Server and Visualization}

\begin{figure}[b] \centering
\begin{tabular}{cc}
\includegraphics[width=0.48\textwidth]{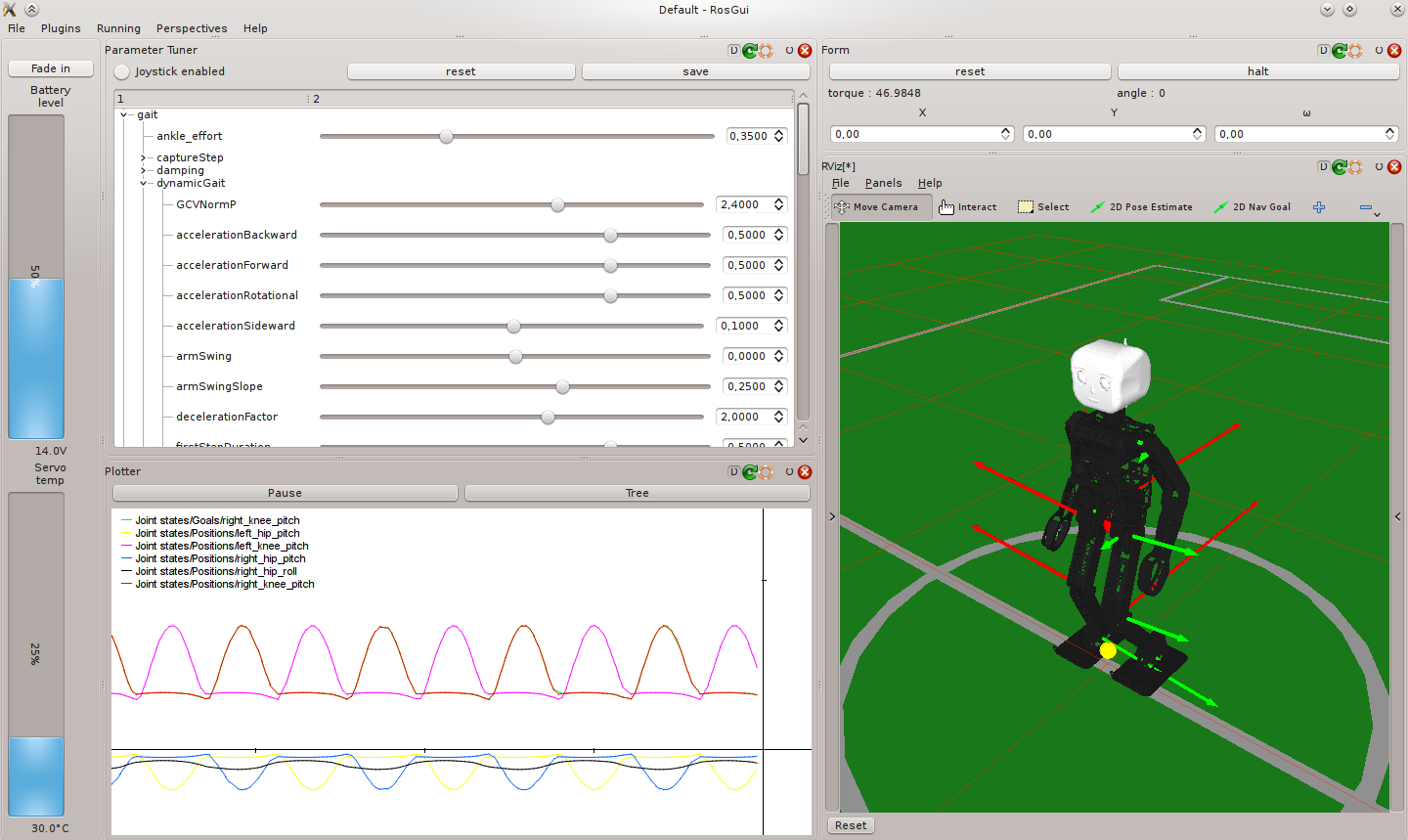}~~ &
\includegraphics[width=0.48\textwidth]{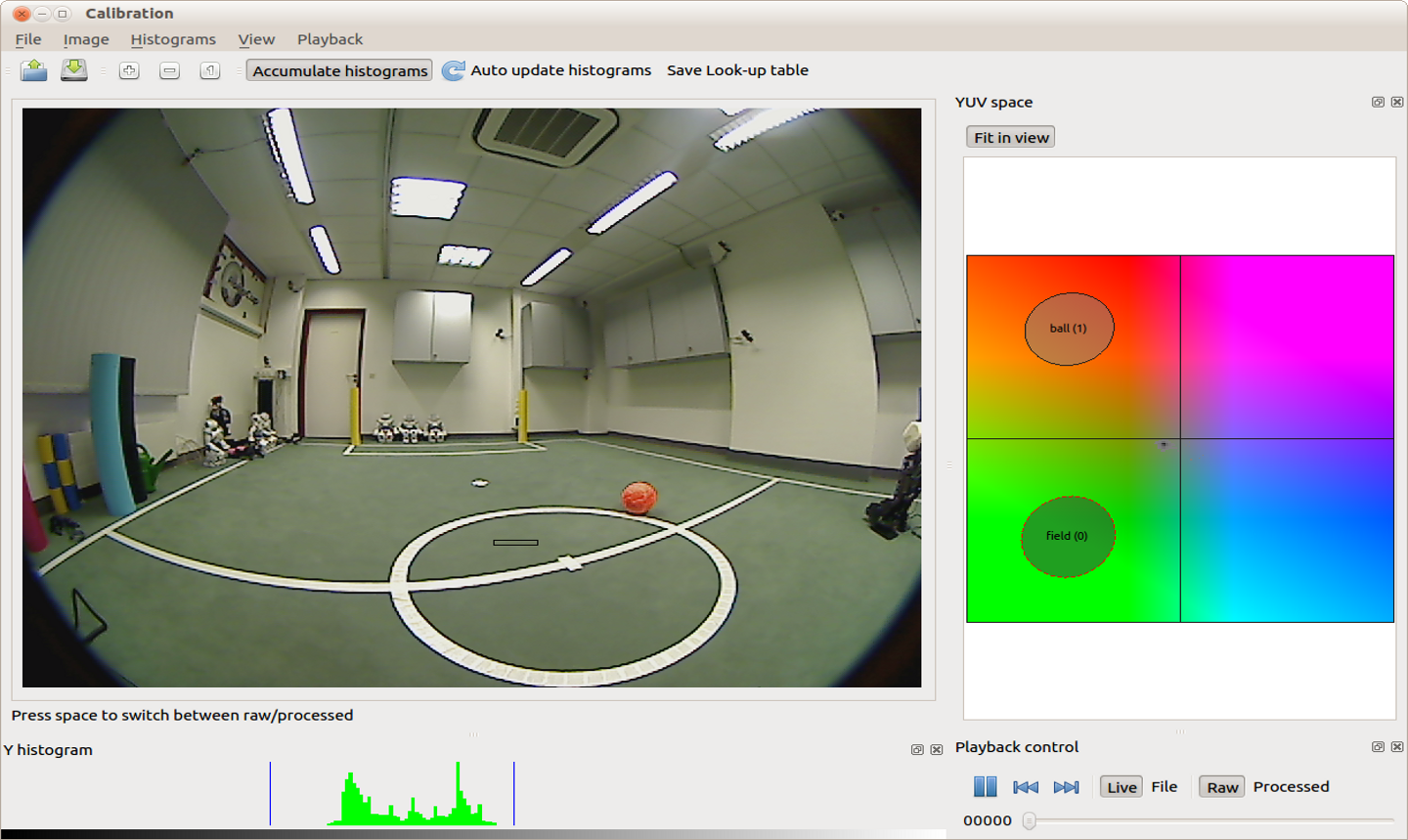}  \\
(a) & (b) \vspace*{-3ex}\\
\end{tabular}
\caption{NimbRo-OP visualization tools. a) 3D model of the robot, 2D plots 
indicating joint positions, gait parameter tuning tool and battery status notifier. b) Color calibration 
tool for generating YUV color look-up tables.}
\label{fig-visualization}
\end{figure}

A significant part of our development are the visualization and parameter tuning
tools. For development and debugging, it is important to be able to capture and inspect the information flow through the
system. ROS offers two
fundamental software packages for this purpose: RVIZ and RQT. The former
provides a 3D visualization of the robot and its environment. The latter permits
the visualization of 2D data such as plots and images. 
Fig.~\ref{fig-visualization} shows two examples of visualization
tools that we implemented.

The software
modules that we develop are highly configurable and, hence, contain many parameters.
In order to allow for and keep track of parameter changes, we developed a configuration server. 
It manages parameters in a hierarchical structure and notifies subscribed nodes about changes.
This server also makes robot configurations persistent for future use. 

\subsection{High-precision Timer}
Some of the previously mentioned procedures
require execution with high-precision periodic timing.  In order to ensure this, the Linux native 
\textit{timerfd} interface is being used. Furthermore, the communication thread uses the realtime
FIFO scheduler instead of the non-realtime default scheduler.
The entire motion layer and hardware communications including sending servo commands, position
feedback and IMU readings are executed every $8$\,ms.
\subsection{Model-based Feed-forward Position Control}

Basis for our current actuator interface is a servo model 
that facilitates compliant robot motion~\cite{Schwarz_2013}. This model is used for feed-forward position
control to compensate gravity and dynamic forces known in advance. The model parameters are 
identified using Iterative Learning Control. Our approach tracks desired trajectories with low control gains,
yielding smooth, energy efficient motions that comply to external disturbances. 

\subsection{Soccer Vision}
\label{sec:vision}

\begin{figure}[t]
\centering
\begin{minipage}{0.4\textwidth}
\setlength{\fboxsep}{0pt}
\fbox{\includegraphics[width=0.45\textwidth]{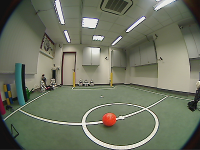}}~
\fbox{\includegraphics[width=0.45\textwidth]{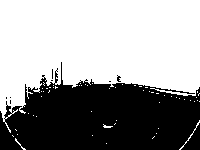}}\\  
\fbox{\includegraphics[width=0.45\textwidth]{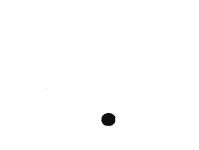}}~
\fbox{\includegraphics[width=0.45\textwidth]{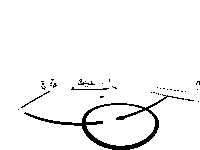}}\\
(a)\\
\end{minipage}
\begin{minipage}{0.55\textwidth}\hspace*{-3ex}
(b)~\includegraphics[width=1.0\textwidth]{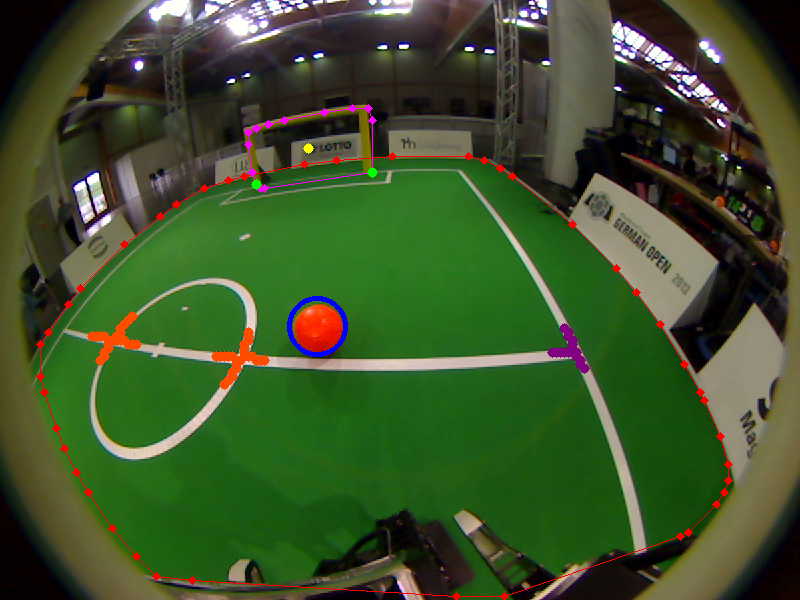}
\end{minipage}
\caption{Soccer vision. (a) Color segmentation (orange, green, white); (b) Identification of field, ball, goal, and line crossings (T and X).}
\label{fig-vision}
\end{figure}

Visual perception of the game situation is a necessary prerequisite for playing soccer.
The robot camera supports resolutions up to 1600$\times$1200. We process $800\times600$ YUYV images to achieve a frame rate above 24\,fps.
We classify all pixels by color and create lower-resolution images for the individual color classes (see Fig.\ref{fig-vision}(a)). 
Using smaller, color classified images facilitates the fast and simple detection of ball, goals, obstacles, field lines, and field line crossings.
Fig.~\ref{fig-vision}(b) shows example detections.
Please refer to Schulz and Behnke~\cite{Schulz:lines} for more detail.

\subsection{Behavior Control}

For initial testing, we implemented a central-pattern generated gait, based on the gait engine of earlier NimbRo robots~\cite{Behnke_2006}.
We also implemented a simple ball approach behavior that makes the robot walk to the ball while simultaneously aiming at the goal~\cite{Stueckler_2008}.
The robot executes kicking and getting-up~\cite{Stueckler_2006} motions based on trajectories designed in a motion editor that we developed.

\section{Impact}
\label{chap:impact}

The initial release triggered quite some interest from the RoboCup community and the general public.
Several RoboCup teams expressed interest to acquire the robot from the University of Bonn or to
manufacture it in their own lab using the released CAD files.
There was even a vivid interest from resellers.
Many news outlets reported on the release of the NimbRo-OP robot, which yielded more than 
32,000 views on its release video. 
The first software release of our robot was demonstrated at the IROS 2012 and Humanoids 2012 conferences.
The newer ROS-based software was demonstrated at ICRA 2013 and RoboCup German Open 2013, where NimbRo-OP received the HARTING Open-Source Award.

Our robot contributed to an increased interest in the Humanoid TeenSize class. 
For RoboCup 2013, the number of qualified teams in the TeenSize class reached an all-time high of six.

\section{Conclusions}
\label{chap:conclusion}

NimbRo-OP is a bipedal TeenSize robot prototype for research. It is
expected that it will be of particular interest to those research groups that
want to work with a larger robotic platform able to act in everyday
environments. The robot has comparably low hardware costs, and it is easy to assemble and
operate. Moreover, if necessary, the robot can be repaired by non-specialists
and can be modified to suit other applications. Both hardware and software
have been released open-source to support low-cost manufacturing and
community-based improvements. The robot is equipped with high torque actuators, 
a dual-core PC for processing power, a wide-angle camera, and a software package that
implements numerous fundamental robot skills.

Further development is planned in order to improve the hardware and software.
With respect to the hardware, special attention will be paid to appearance,
durability, impact resistance and easy maintenance.
The NimbRo-OP ROS framework, which is still a work in progress,
will receive additional functionality. For example, a self-localization node 
that uses information from the vision detection and the robot state to feed a
particle filter model \cite{Schulz:lines} and a simulator, where the
user will be able to test behaviors and new algorithms before deployment on the
real robot, will be among the next steps.

\subsection*{Acknowledgements}\vspace*{-1ex}

This work has been partially funded by grant BE 2556/6 of German Research Foundation
(DFG). We also acknowledge support of Robotis Inc. and RoboCup Federation for the first robot prototype.

\bibliographystyle{unsrt}
\bibliography{ms}

\end{document}